# Critical appraisal of artificial intelligence for rare-event recognition: principles and pharmacovigilance case studies

G. Niklas Norén, Eva-Lisa Meldau, Johan Ellenius

Uppsala Monitoring Centre, Uppsala, Sweden

niklas.noren@who-umc.org

## Abstract

Many high-stakes AI applications target low-prevalence events, where apparent accuracy can conceal limited real-world value. Relevant AI models range from expert-defined rules and traditional machine learning to generative LLMs constrained for classification. We outline key considerations for critical appraisal of AI in rare-event recognition, including problem framing and test set design, prevalence-aware statistical evaluation, robustness assessment, and integration into human workflows. In addition, we propose an approach to structured case-level examination (SCLE), to complement statistical performance evaluation, and a comprehensive checklist to guide procurement or development of AI models for rare-event recognition. We instantiate the framework in pharmacovigilance, drawing on three studies: rule-based retrieval of pregnancy-related reports; duplicate detection combining machine learning with probabilistic record linkage; and automated redaction of person names using an LLM. We highlight pitfalls specific to the rare-event setting including optimism from unrealistic class balance and lack of difficult positive controls in test sets - and show how cost-sensitive targets align model performance with operational value. While grounded in pharmacovigilance practice, the principles generalize to domains where positives are scarce and error costs may be asymmetric.

## Key points

- Rare-event recognition poses unique challenges for AI evaluation: apparent accuracy can be misleading when prevalence is low, and test sets may fail to capture difficult positives

- We outline key considerations in appraisal of AI models for rare event-recognition, including prevalence-aware statistical evaluation, robustness assessment, and integration into human workflows
- Structured Case-Level Examination (SCLE) is introduced as a novel complement to contextualize statistical performance evaluation, generating insights from human review of false positives, false negatives, and correct classifications
- A comprehensive checklist is provided to guide procurement or development of AI systems for rare-event recognition

# Introduction

With the rapid evolution in performance and versatility of artificial intelligence systems, their use is becoming more widespread and valuable (1,2). The ability to appraise such systems is important for professionals and decision-makers to invest their resources wisely. A failure to effectively evaluate AI systems may lead to time and effort being wasted on systems that do not deliver what they promise or even harm individuals or the public good. Whereas human operators also make mistakes and vary in their skillfulness, the impact of a single flawed human operator is typically limited, whereas a single flawed AI system can be deployed at scale with widespread impact. On the other hand, unwarranted skepticism may mean organizations and individuals miss opportunities and benefits that AI systems could bring.

In general, AI may yield improvements of the following general nature:

- **Efficiency**: performing tasks that humans would otherwise do - but faster, with less effort
- **Quality**: performing tasks that humans would otherwise do - but more accurately, consistently
- **Capability**: performing tasks that otherwise would not get done

The distinction between quality / efficiency and capability is not always clear, and what constitutes new capability may vary between human operators. For example, an English-to-German machine translation would offer new capability to co-author GNN who does not know German but (possibly) efficiency or quality to co-author ELM who is a German native. Often, the value may not be achieved by the artificial intelligence system alone but by a human-AI team. Then, the aim may be intelligence *augmentation* – to support and enhance human decision-making.

The use of artificial intelligence as a term to describe certain computer science applications is growing in popularity again, but interpretations and delineations vary.

Whereas its colloquial use may imply a certain degree of autonomy and versatility, technical definitions tend to be more inclusive. For example, Aronson proposes the following definition of artificial intelligence, after careful consideration of existing definitions and their limitations (3):

> ***artificial intelligence*** *n. a branch of computer science that involves the ability of a machine, typically a computer, to emulate specific aspects of human behaviour and to deal with tasks that are normally regarded as primarily proceeding from human cerebral activity.*

and the OECD define an *AI system* as (adopted also in the European Union's AI Act) (4):

> ***AI system:*** *An AI system is a machine-based system that, for explicit or implicit objectives, infers, from the input it receives, how to generate outputs such as predictions, content, recommendations, or decisions that can influence physical or virtual environments. Different AI systems vary in their levels of autonomy and adaptiveness after deployment.*

These definitions would include expert systems that have been explicitly programmed to perform specific tasks, as well as any application of machine learning where computational methods infer and optimize behavior based on training data. The former can range from simple heuristics to highly complex and capable systems, like IBM's Deep Blue chess-playing computer, which combined massively parallel search with expert-designed evaluation functions and human opening/endgame knowledge (5). The latter includes both lower-dimensional data-driven linear or rule-based models and large language or image models using deep neural networks with trillions of parameters (and counting).

Whereas AI systems can perform a variety of tasks, our focus here is on AI models for recognizing rare events (6,7) to support organizational workflows or to enrich datasets for downstream analysis. Examples include spam or phishing detection (where messages are flagged for review are in minority) (8), fraud detection (where rare suspicious transactions are escalated for audit) (9), safety monitoring in aviation or cybersecurity, and computational phenotyping as a basis for e.g. epidemiological research (10). In contrast, applications such as clinical diagnosis, where predictions directly inform individual patient care, fall outside our scope. In pharmacovigilance, rare-event recognition plays a role both in day-to-day case processing and data management — for example, by flagging suspected duplicates or reports that warrant expert review — and as a foundation for subsequent aggregate analyses in signal management (11,12). For this reason, we draw on use cases from pharmacovigilance to illustrate and ground the principles proposed in this paper.

To determine the appropriate scope and approach for appraising an AI system the following qualities may be relevant: **ambiguity of task** (is the correct classification of events clear to human operators); **opacity** (can humans understand the factors and inner logic based on which the *AI system* arrives at its classification); **adaptiveness** (can the *AI system* adjust its classifications to changing conditions or feedback regarding its performance); **scope** (how wide a variety of events can the AI system recognize and under what conditions can it operate); **autonomy** (what level of human oversight and control is the AI system subject to). With no ambiguity, standard computer validation should suffice, whereas with higher ambiguity, statistical performance evaluation will be required, and more care must be exercised in defining reference standards. With increasing opacity, expectations on transparency regarding the nature of the AI model, its training, and performance evaluation should increase (13,14). With higher adaptiveness, expectations on continual performance monitoring increase. However, any AI model can be vulnerable to data drift (15–17), and when AI models with broader scope are applied to new types of events or operating conditions, additional performance evaluation may be required. With higher autonomy, measures to ensure the validity, robustness and safe use of AI must be more extensive, according to the risk-based approach.

A wide variety of AI models may be relevant for rare-event recognition, including expert-defined rule-based methods, traditional machine learning methods (e.g. SVMs or gradient boosted trees), fine-tuned LLMs with a classification layer (e.g. BERT (18)) generative LLMs (e.g. GPT4 (19)) with constrained output (e.g. through prompting or post-processing of their textual output). See Figure 1.

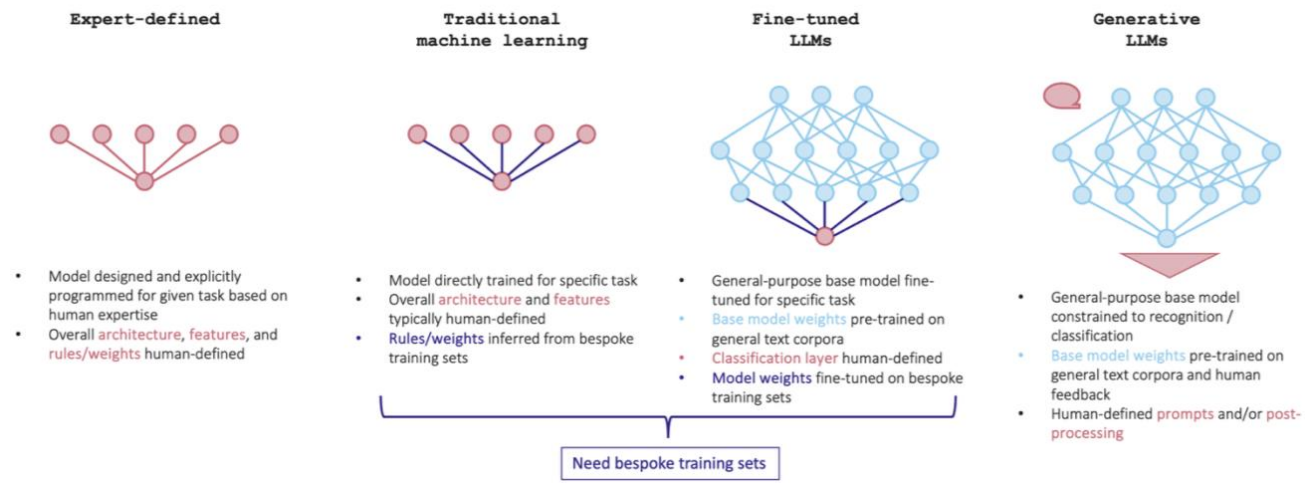

*Figure 1*. Different types of AI models for rare-event recognition. *Red* graphical elements and text highlight model components defined based on human expert input, e.g. 'architecture'; *dark blue* elements highlight components inferred from bespoke training sets for the task at hand, e.g. 'Model weights'; *light blue* elements highlight components inferred during general purpose pre-training ('Base model weights').

To benefit from the opportunities that AI may bring while avoiding the pitfalls, professionals and decision-makers must acquire a basic capability to critically review and appraise proposed artificial intelligence systems. Validity and robustness are fundamental requirements on trustworthy AI, and the ability to assess AI model performance evaluation results is crucial for those who develop or procure AI systems.

The aims of this paper are to demonstrate that those requiring rare-event recognition must be able to critically appraise artificial intelligence systems, and to equip them with practical capabilities to assess performance evaluations. Specifically, we propose a structured approach to case-level examination to complement and contextualize statistical performance evaluation, and a comprehensive checklist to support procurement or development of AI models for rare event-recognition. This should be viewed as a complement to more general guidance on the topic (20,21).

# Examples used for illustration

Throughout this article, we will refer to three recently published evaluations of artificial intelligence models in pharmacovigilance for illustration: an expert-defined rule-based method for recognition of adverse event reports related to pregnancy (22), a new version of a machine learning-based model for improved duplicate detection in large collections of adverse event reports (23), and a fine-tuned deep neural network for automated redaction of person names in case narratives (24). The methods are different in nature but face many of the same challenges during evaluation. For an overview of intended use and test sets for our three use cases, refer to the table in Appendix A.

## Rule-based retrieval of pregnancy reports

Sandberg et al (22) developed a rule-based method to identify adverse event reports involving exposure to medicinal products during pregnancy, affecting either the pregnant individual or the prenatally exposed child. The method begins by preprocessing reports to align with the ICH E2B(R2) data model, including conversion of adverse events to the MedDRA terminology when necessary. It then applies two sets of rules: one to exclude reports that are highly unlikely to involve pregnancy, and another to identify those that are likely relevant. This method is available as an optional search filter in VigiLyze, a tool for members of the WHO Programme for International Drug Monitoring to analyze reports in VigiBase, the WHO database of adverse event reports for medicines and vaccines.

## Duplicate detection with SVM and statistical record linkage

Barrett et al (23) combined statistical record linkage with traditional machine learning to improve duplicate detection in adverse event reporting systems. Duplicate reports are unlinked reports referring to the same adverse event in a given patient that can distort analyses and waste reviewer resources if not identified and removed. The new approach combines a Support Vector Machine (SVM) classifier with principles from statistical record linkage. It compares report pairs on multiple features, some defined by experts and others derived from statistical modelling. The method rewards unusual similarities (e.g. two patients sharing the same rare adverse event) and penalizes unlikely mismatches. The improved model is currently being implemented to support workflows where suspected duplicates are flagged for human review or automated preprocessing, where duplicates are removed before statistical signal detection. By combining machine learning with structured statistical logic, the method aims to improve performance compared with earlier versions of vigiMatch, while remaining transparent enough to align with regulatory expectations.

## Redaction of person names with fine-tuned LLM

Meldau et al (24) fine-tuned an LLM, in this case BERT, with a classification layer to predict which token (word or other sequence of characters) in a case narrative text represents a person name. The BERT model had been pre-trained on English text including books and Wikipedia texts and was fine-tuned on combined data from a public de-identification challenge dataset called 2014 i2b2/UTHealth Corpus and case narratives from UK Yellow Card data. It is intended to either fully automate redaction or support a human redacting case narratives, i.e. to highlight parts of the text that should be masked or replaced by a placeholder. This is intended for an early step of case processing soon after it arrives at the pharmacovigilance organization.

# Test set-construction for rare events

In rare-event recognition, we may refer to those data points that we want to retrieve as *positive controls* and those that we do not want to retrieve as *negative controls*. We use this terminology throughout the description below.

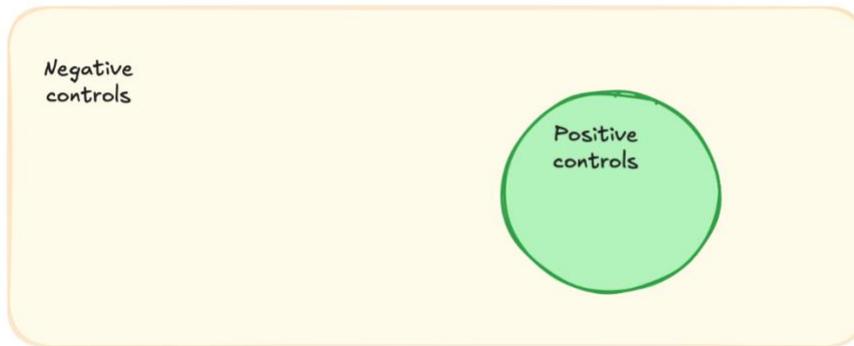

The nature of a test set should align with the desired deployment domain. For example, in evaluating methods for pharmacovigilance signal detection, historical safety signals would typically be a better choice of positive controls than well-known, already labelled adverse drug reactions because their reporting patterns differ (25). Similarly, if an AI model for recognizing adverse events in free text is intended for broad use across collections of adverse event reports, the test set should include reports related to a broad range of drugs and adverse events, from both patients and health professionals, etc.

Sometimes, boundaries between positive and negative controls are not clear-cut, which yields additional sources of ambiguity. For example, different organizations may have different requirements on how strong the conviction should be that two individual case reports refer to the same event for them to be classified as suspected duplicates. Similar considerations apply in defining negative controls: one may use all data points which do not meet the criteria to qualify as positive controls or establish a margin-of-error in which data points that are difficult to classify are set aside and do not count as either positive or negative controls. For example, because of ambiguity in the presentation and encoding of adverse events, one may choose to exclude from the negative controls any terms adjacent to a positive control (e.g. MedDRA Preferred Terms in the same High-Level Term group as a positive control).

For transparency, the nature of test sets should be clearly specified and design choices made during their development documented and communicated (13). It may be relevant to present high-level descriptive statistics for positive and negative controls. This is important especially when a method may be used for variations of the intended use case in the future. The development of annotation guidelines outlining the principles based on which data points are classified as positive or negative controls in creating test sets sometimes improves quality and consistency of human processing by harmonizing and making explicit decision processes that were previously implicit and variable within an organization.

For rare events, straight random samples of data points will contain a low proportion of positive controls, and it may be impossible or too costly in time and effort to annotate a sufficient random sample to obtain enough positive controls. On the other hand, if enrichment strategies are used to increase the proportion of positive controls in the test set, this may yield misleading performance estimates, if not appropriately accounted for in the analysis, as discussed below.

We focus here on test sets, because our interest is primarily in performance evaluation. Reference sets are typically divided into training, validation, and test splits, where the former two are used for model training, fine-tuning, and hyper-parameter tuning. Many considerations in this section are relevant also during model training and validation. However, more liberty can be allowed in those phases, and some approaches that would not be acceptable in test-set construction, can be applied, with care. For example, in the fine-tuning the BERT model for the UMC redaction of narratives method, we took advantage of the tendency of person names to cluster within narratives, by directing our annotators toward narratives that a simpler classifier had suggested contained at least one person name. Similarly, the training and validation data for the SVM model in vigiMatch duplicate detection was actively enriched with positive controls (some of which were identified by earlier versions of the method under development). Each of these practices could have resulted in suboptimal performance or generalization of the trained models, and it is critical to scrutinize fitted models and their interim performance evaluation results on validation data during development to recognize and course correct when this happens. As an example, during the development of vigiMatch, inspection of an interim SVM model showed that it would *penalize* two reports if they came from the *same* country, which is counter intuitive. The reason, it turned out, was an over-representation of reports from one specific country which had had an excessive rate of false positives for an earlier version of the model. This had led many non-duplicate report pairs from this country to be annotated and included in our training set, which could then be addressed.

## Prevalence-aware statistical performance evaluation

For rare event-recognition where we cannot require or expect perfect performance, and where the task itself may be ambiguous (i.e. humans may not be able to classify all data points with certainty), evaluation of AI models will typically be statistical.

The basis for statistical performance evaluation for a binary classification task is a cross-classification of data points comparing the predictions of an AI model with the annotations in the test set. A positive control predicted positive by the AI model is referred to as a True positive (TP). A positive control predicted negative by the AI model is referred to as a False negative (FN). A negative control predicted positive by the AI model is referred to as a False positive (FP) and a negative control predicted negative is referred to as a True negative (TN).

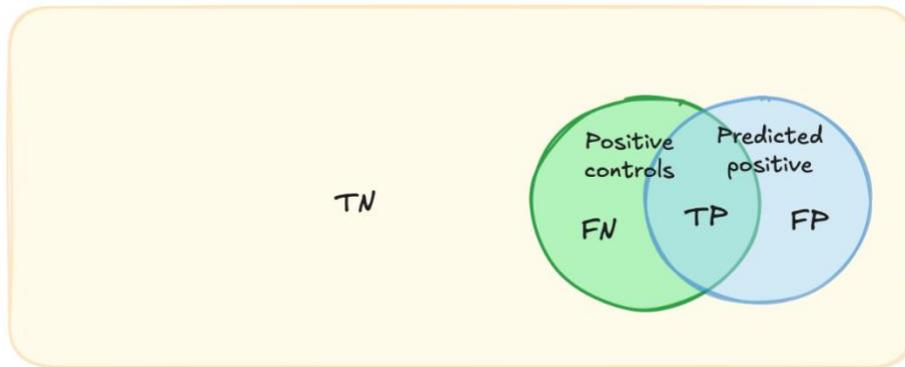

*Figure 2*. Basis for statistical performance evaluation in a binary classification task. Any data point outside the blue ellipse is Predicted negative, and any point outside the green ellipse is a Negative control. TP = True positive; FP = False positive; FN = False negative; TN = True negative.

The above cross-classification presumes that, for AI models with continuous output, a decision threshold has been set. The selection of the threshold should account for the relative costs of different types of errors (false positives vs false negatives) and may be guided by domain experts' input on acceptable performance levels. Detailed performance analyses may assess a range of decision thresholds as discussed below. The relative costs of errors may vary with the intended use of a given method, and so may the relevant decision thresholds. For example, false positives in duplicate detection cost time and effort (and may lead to alert fatigue) if forwarded to human operators for review but could lead to missed/delayed signals if suspected duplicates are removed prior to statistical signal detection.

In rare-event recognition with a focus on organizational impact and downstream analyses, recall, precision - and in certain circumstances specificity - are core performance metrics. Together, they account for false negatives (recall/sensitivity) and false positives (precision/positive predictive value or specificity). Negative predictive value (the probability that a predicted negative is correct) like recall/sensitivity focuses on false negatives but asks how much an individual negative prediction can be trusted. It is important in medical diagnostic, for example, but less relevant for the applications in focus here. However, it does provide useful information on the purity of the set of predicted negatives for downstream analyses, which may inform to what extent misclassified positive controls may bias subsequent analyses. As discussed below, precision provides similar information regarding misclassification of negative controls. For a complementary discussion of pitfalls in AI performance evaluation, with a slightly different scope, see e.g. Hicks et al (2022) (26).

## Recall

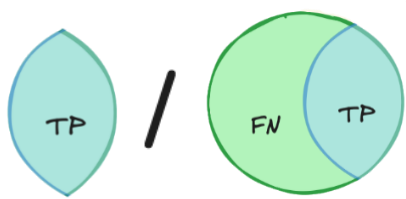

**Recall** is the proportion of positive controls that are predicted positive by the AI model: $\frac{TP}{TP+FN}$. It measures how many of the data points that we are interested in that are correctly classified – or recalled. *Sensitivity* and *True positive* rate are synonyms.

Recall is relevant in applications where it is important to identify events of interest as completely as possible, and false negatives are costly – for example, if unrecognized events mean appropriate actions are not taken or are delayed. For downstream analyses, low recall means that not all true events are recognized which may reduce statistical power.

To compute recall requires a representative set of positive controls. In rare event settings, annotation of random samples can be costly, and if enrichment heuristics are used to increase the proportion of positive controls in the test set, recall may be over-estimated if positive controls that are harder to recognize for the AI system are also less likely be found with the heuristic. If positive controls are not easily recognized / classified by human operators upon review, misclassification can lead to optimistic recall estimates (if AI struggles with the same cases).

| |
|---|
| **Example: VigiBase pregnancy algorithm** <br> - Published study annotated random sample of reports, after restriction based on patient age and sex (i.e. no active enrichment with positive controls). Demonstrated that the impact of the restriction by age and sex on estimates of recall (by potentially excluding hard positive controls) was negligible |
| **Example: vigiMatch duplicate detection** <br> - Used sets of known duplicates already identified by four different national regulators to study recall (i.e. no new annotations and therefore no enrichment) <br> - Acknowledged that these test sets may fail to include true duplicates not recognized as such by human operators (either for lack of sufficient information on individual reports or because there exist too many pairs for humans to assess all possible pairs) which may lead to optimistic recall estimates. |
| **Example: UMC redaction method** <br> - Annotated random sample of 5,042 case narratives identifying 179 NAME tokens in 71 narratives <br> - Conservative classification of edge case as NAMEs when in doubt, which should lead to conservative (possibly pessimistic) recall estimates. |

## Precision

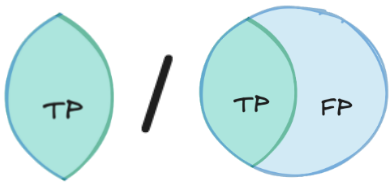

**Precision** is the proportion of predicted positives that are positive controls: $\frac{TP}{TP+FP}$. It measures how many of the data points that are classified as of interest by the AI model that are correct according to the reference standard. *Positive predictive value* (PPV) is a synonym.

To compute precision requires a representative set of data points that are predicted positive by the AI model of interest. Precision is relevant in applications where the proportion of false positives among predicted positives matters – for example when there are humans in the loop and each false positive requires human effort to process. Low precision then reduces efficiency because of time is spent processing false positives and may lead to alert fatigue. For downstream analyses, precision reflects the purity of the set of predicted events, and the risk of misclassification bias from incorrectly flagged events.

Importantly, the estimated precision is highly dependent on the prevalence of positive controls in the test set, and if test sets have been enriched with positive controls, naive test set precision estimates will be optimistic and not reflect real-world performance. Note that the expected estimated precision of random guessing equals the prevalence of positive controls (so for a test set balanced on positive and negative controls, 50% precision is the baseline).

Precision is in principle straightforward and relatively cheap (in time and effort) to estimate for a given AI model by applying it to a random sample of data points and annotating all predicted positives. However, such *model-specific precision test sets* will need to be replaced or updated if the AI model is changed or another model is considered. As such, they are not suitable to reuse for benchmarking.

'Precision@k' measures precision for the highest decision threshold which results in k predicted positives. It can be a relevant when existing operating points are of that nature, for example when teams operate under a fixed review budget. It can also enable comparative analyses between different models or across time, when decision thresholds are difficult to calibrate or unstable.

> **Example: VigiBase pregnancy algorithm**
> - A random sample of 30,000 reports were retrieved from the full dataset, and all 448 predicted positives by the algorithm or its benchmark within this sample were annotated. This sample size was chosen based on simulations showing it would provide sufficient statistical power (80% at a 5% significance level) to detect a

| |
|---|
| true relative difference in precision of at least 10% between the algorithm and its benchmark. |
| **Example: vigiMatch duplicate detection**<br>- Model-specific precision tests for method of interest and its benchmark based on ~100 predicted positives were annotated; careful re-use of random report pair sequences ensured maximal overlap between the two model-specific precision tests, reducing the number of required annotations.<br>- Too low prevalence of negative controls in training and validation sets during early model development gave models with unacceptably low precision in real-world setting - revealed by model-specific precision tests during model training and validation (unpublished results) |
| **Example: UMC redaction of narratives method**<br>- Precision computed on the completely annotated randomly sampled test set described above which contained 263,272 NON-NAME negative control tokens along with the 179 positive control NAME tokens. Whereas precision may provide a sense of how end users will perceive the performance of the method (what proportion of redacted tokens correspond to NAMEs), specificity was considered the most relevant measure of impact from false positives as discussed below. |

## Specificity

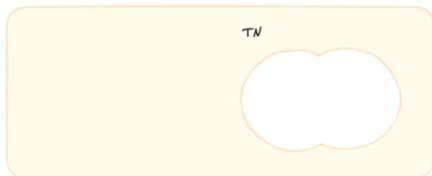

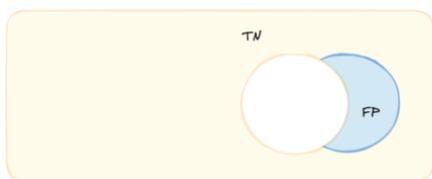

**Specificity** is the proportion of negative controls that are predicted negative by the AI model: $\frac{TN}{TN+FP}$. It measures how many of the data points that we are **not** interested in, are correctly classified as negative. *True negative rate* is a synonym.

When some action is taken for every predicted positive, specificity may be more relevant than precision to assess the impact of false positives - or at least serve as a useful complement. One advantage is its independence of the prevalence of positive controls in the test set. Based on estimates of sensitivity and specificity, precision can be computed for different levels of (assumed) prevalence of positive controls, using Bayes theorem.

However, to estimate specificity, requires a representative (and large enough) set of negative controls and since negative controls far outnumber positive controls in rare-event recognition, specificity needs to come very close to 1 to achieve acceptable precision in many applications. This in turn requires a very large number of negative controls which can be impossible, or at least very costly to obtain by annotating a random sample. This limits the usefulness of specificity as a performance metric in many rare-event recognition applications.

> **Example: vigiMatch duplicate detection**
> - Specificity *not* reported as part of the study. Specificity at report *pair* level by necessity extremely close to 1 and practically impossible to estimate. However, specificity at the report-level (how many non-duplicates are incorrectly flagged by the method) could be valuable in assessing the collateral impact of removing suspected duplicates prior to subsequent analyses and may be considered in the future.
>
> **Example: UMC redaction of narratives method**
> - False positive rate (= 1-specificity) estimated to 0.05% used as a measure of the impact on the utility of the redacted narrative; the impact being high if many NON-NAME tokens were removed, making it harder to read and potentially removing clinically relevant information. Note that in this case the specificity of 99.95% resulted in a precision of only 55%, because of the low prevalence of positive controls. Using Bayes formula, we see that if specificity had instead been 98% with prevalence and recall unchanged, precision would drop to 3%. In other words, almost all redacted tokens would be NON-NAMEs, even though the nominal specificity may seem high.

## Composite performance metrics

**F1-score** (sometimes just F-score) aggregates the estimated recall and precision for a specific threshold into a single measure of predictive performance. It is computed as the harmonic mean (a special type of average) between precision (*p*) and recall (*r*):

$$F1 = \frac{2 \cdot p \cdot r}{p + r}$$

It inherits the possible limitations of precision and recall described above. Moreover, it assumes equal costs of false positives and false negatives, which is appropriate only in special casees.  Variations of the F-score such as the F2-score make different assumptions of error costs but are not straightforward to interpret.

**Precision-recall curves** are a tool to display threshold-independent performance by plotting precision and recall for different values of the decision threshold. They can be valuable to describe performance at different operating points, when precision and recall are the relevant metrics. However, they suffer from the same limitations and risks of misleading results as their component metrics, when test sets are enriched with positive controls.

**Receiver-Operating Characteristics (ROC) curves** are another tool to display threshold-independent performance by plotting recall (sensitivity) against (1 – specificity) in a 2d graph for different values of the decision threshold. Performance better than chance is above the diagonal, and the closer the curve comes to the top left corner, the better the performance. Overall performance is often measured as the Area Under the Curve of the ROC curve (AUC or AUROC), which corresponds to the probability that a randomly selected positive control is ranked above a randomly selected negative control by the assessed method. AUC values greater than 0.5 are better than chance.

ROC curves inherit the limitations of recall/sensitivity and specificity highlighted above. Whereas specificity can sometimes be relevant in rare event settings, only very limited portions of the ROC curve are of interest for rare events which limits their usefulness. Because overall measures such as AUC are based on the entire range of specificity values from 0 to 1, they will be dominated by portions of the decision curve of no consequence in a rare-event setting.

| |
|---|
| **Example: VigiBase pregnancy algorithm** |
| - The pregnancy algorithm does not rely on a tunable decision threshold, as it consists of a fixed set of rules. Potential improvements in recall could be achieved by modifying the rule set, specifically, by reducing the number of exclusion criteria used to rule out cases or expanding the inclusion criteria used to identify pregnancy-related reports. |
| **Example: vigiMatch duplicate detection** |
| - No composite performance metrics like F-scores presented as part of study, since relative costs of false positives and negatives considered to vary with the specific use case
- Threshold-independent analyses such as precision-recall graphs may have helped assess impact of decision threshold but were not obtained as part of the study.
- Specificity at report pair-level extremely close to 1 and ROC curves considered not relevant or informative. |
| **Example: UMC redaction of narratives method** |
| - F1-scores computed to allow comparison to other studies and to summarize precision/recall into a single score. However, recall was seen as most important (and method optimized for recall during development).
- Precision-recall graphs used during model development to choose threshold in the BERT classification layer.
- Specificity 0.995 for selected threshold and ROC curves considered not relevant or informative. |

## Robustness analyses

Most classifiers for rare-event recognition are deterministic. Once trained and fine-tuned (if machine learning-based), they will generate the same output for any given input. However, classifiers based on generative LLMs include stochastic components (sometime controlled with hyper-parameters like 'temperature') and may classify the same data point differently in repeated execution.

**Stability** is a general performance metric reflecting the tendency of a non-deterministic AI system to generate similar or the same output under limited (or no) perturbations of the underlying data. There is not a single stability metric. Rather, stability must be defined with a specific application in mind. For example, stability in rare-event recognition may be assessed as the proportion of data points assigned the same labels in repeated execution of a non-deterministic AI model

More broadly, the robustness of training or fine-tuning a machine learning model can be assessed by considering the variability in fitted models (or their individual predictions or overall performance) when trained on different subsets of training data – e.g. folds of a cross-validation or bootstrap samples.

A further aspect of robustness concerns performance on specific subsets of the data. Global performance metrics may mask important heterogeneity: a classifier may perform well overall, while failing systematically for subgroups of cases defined by e.g. their features, source, demographics, or time period. Subset-specific analyses are relevant to ensure both fairness & equity and validity & robustness. From a fairness perspective, it is important not to under-serve or explicitly bias against certain subgroups. For downstream analyses, differential performance in recognizing an event of interest across subsets may introduce bias. Robustness analysis should thus consider stratified performance estimates and, where feasible, assess whether the model's errors are randomly distributed within subsets or cluster around some data characteristics. Such analyses do not necessarily require new metrics, but rather a systematic breakdown of standard measures (e.g. precision, recall, specificity) across relevant subsets.

Transparency of robustness analyses helps support trustworthy use of AI by clarifying the boundaries of reliable performance (13). If limitations are identified, these can guide mitigations—such as targeted threshold adjustments, abstain/triage strategies for vulnerable subsets, or prioritization of further training data collection.

> **Example: VigiBase pregnancy algorithm**
> - The method's recall increased from 75% overall to 91% when applied specifically to reports adhering to the ICH E2B format. This subset-specific analysis reveals

| |
|---|
| how structural differences in input data can significantly affect performance, underscoring the importance of stratified evaluation. |
| **Example: vigiMatch duplicate detection**<br>- To assess performance for individual countries, small-scale model-specific precision tests were performed for two African and one European countries; this was done because the benchmark-method had been found to perform less well in countries in individual countries with adverse event and drug distributions which differ substantially from the global pattern. |
| **Example: UMC redaction of narratives method**<br>- Sensitivity analyses focused on the one false negative corresponding to a full person name, which was of Indian origin (to assess possible issues with fairness & equity). Data manipulation experiments showed that the method was able to recall other names (of various origins) when inserted in the narrative of interest (except, strangely, the first name in 'John Smith') and that the name of interest was appropriately redacted when inserted in one of the other narratives in the study. This indicated that the failure here was due to some interaction between the name and the narrative. |

## Benchmarks

Comparison to relevant benchmark methods provides an important point of reference when evaluating novel AI models, especially for bespoke test sets, whose nature, scope, strengths, and limitations may not be easy to assess for outsiders. For more complex benchmark methods, appropriate hyper parameter-tuning should be ensured to obtain relevant results.

Where available, public benchmark test sets offer additional value by enabling standardized comparisons across studies. In specialized applications such resources may be less common, and benchmarks from adjacent domains could be considered, though such comparisons require careful interpretation given differences in context and data characteristics. Unfortunately, benchmark methods and test sets remain scarce in specialized domains, like pharmacovigilance, limiting opportunities for standardized comparison.

| |
|---|
| **Example: VigiBase pregnancy algorithm**<br>- **Benchmark method** Standardised MedDRA Query (SMQ) for pregnancy-related terms<br>- **Benchmark reference set**: Not available |
| **Example: vigiMatch duplicate detection**<br>- **Benchmark method**: Earlier version of vigiMatch, representing current state-of-the-art (for adverse event reports related to drugs; for vaccine reports, no benchmark method available). |

|  |  |
|---|---|
| - **Benchmark reference set:** Not available | |
| **Example: UMC redaction of narratives method** | |
| - **Benchmark method**: not available for the specific intended use case | |
| - **Benchmark reference set**: Test set from i2b2 challenge (acknowledging that this is not fully aligned with the intended use case and not what the method has been optimized for) and compared to state-of-the-art methods. | |

## Structured case-level examination (SCLE)

We propose a Structured Case-Level Examination (SCLE) complement statistical performance evaluation in critical appraisal of AI models with case-level review of errors and correct classifications. SCLE extends and systematizes a practice we have applied informally in the past, including in the three case studies presented in this paper.

Summary performance metrics go only so far in enabling us to assess and understand the performance of an AI model. Equally important is to inspect representative, concrete examples of an AI model's classifications. Such examples should be analyzed during AI model development, evaluation, and performance monitoring / re-training. They should ideally be communicated to end users and in scientific publications when performance evaluation results are described.

An SCLE should include both correct and incorrect classifications. Examining false positives and false negatives can each give useful insights regarding the strengths and limitations of the AI model and its evaluation and contextualize statistical performance evaluation. For example, if a false negative in redacting case narratives corresponds to a full name preceded by 'Mr', this may undermine end users' trust in the AI system, even if the aggregate recall estimate is excellent. On the other hand, if the false negative is 'AF' and it is hard to know for a human specialist from the surrounding text if these are initials or an abbreviation for *atrial fibrillation*, then the overall precision metric may be viewed as potentially conservative in view of the ambiguity. Review of correctly classified data points may in turn give insights regarding the difficulty of the tasks successfully performed by an AI model. This may be especially important when there is no benchmark method that can provide baseline comparator, and we may not understand from overall performance metrics the difficulty of the task at hand.

For optimal use of human resources, we propose, for rare event-recognition, to examine a stratified random sample of individual i) false positives, ii) false negatives, and iii) true positives. True negatives are the majority case and typically of less interest in in rare-event recognition. If there are subgroups of special interest in the performance evaluation, one may sub-stratify to ensure relevant coverage of these. It may also be relevant to sub-stratify by distance to the decision boundary – for example 'confident' misclassifications can be reviewed as part of sanity checking. Generally, a risk-based approach should determine the relative importance of i), ii), and iii); the corresponding sample sizes; and the stringency of the human review and labelling.

Human review may consider a set of diagnostic tags relevant to the case at hand, in addition to free text notes. For example:

- '**Never events'** (misclassifications that would not be acceptable or could severely undermine trust in AI)

- **Unexpected errors** (may point to opportunities to improve model and/or issues with training set)

- **Input data issue** (surprising quality issues, unexpected nature/scope of data points, human classification based on information not available to AI)

- **Test set issue** (incorrect or ambiguous labels, too high or low granularity compared with intended use)

- **Triviality** (consider a simple metric like non-trivial / trivial / unclear for use in aggregate)

When there is a benchmark method, one may focus the SCLE on data points that are differentially classified by the method of interest and the benchmark, to better understand the nature of differences in performance between the two. In such circumstances, consider oversampling the disagreement cells (Model+, Benchmark– and Model–, Benchmark+). It may also be relevant to include diagnostic tags related to unexpected performance compared with benchmark.

Possible remedial action in response to findings in an SCLE include re-training (changes to training set and/or model), modifying thresholds, data quality improvements, updates to annotation guidelines, and refined communication of performance evaluation.  An over-representation of certain features or types of data among the misclassified may motivate systematic performance evaluations in specific subgroups. During model development SCLE can be instrumental in identifying areas of possible improvement, especially for more opaque methods that do not allow inspection / interpretation.  A high rate of triviality of correct classifications may call for closer scrutiny of the test set.

SCLE is relevant not just pre-deployment but throughout an AI model's lifecycle, for example following re-training or in response to identified data drift (10).

| |
|---|
| **Example: VigiBase pregnancy algorithm**<br>- Case-level review of false negatives identified that pre-processing errors were the most frequent source of these errors. The most critical issue were indications reported in non-English ICD text, which had not been mapped during data preprocessing. Another contributing factor was incomplete mapping of pregnancy-related terms to MedDRA.<br>- Case-level review also confirmed known algorithm-level limitations including the inability to process pregnancy information confined to free-text fields, meaning relevant data was present but not accessible to the rule-based logic. Additionally, some structured fields containing pregnancy-related information were not included in the algorithm's rule set. |
| **Example: vigiMatch duplicate detection**<br>- Benchmark-aware case-level review of true positives, false positives, and false negatives for the vigiMatch model for drugs.<br>- Case-level review of false positives and false negatives for vigiMatch model for vaccines (where no benchmark was available).<br>- Scope limited to handful of cases per classification category; human review subjective in nature without pre-specified diagnostic tags. |
| **Example: UMC redaction of narratives method**<br>- Systematic case-level review of *all* false negatives classifying into diagnostic tags for directly, indirectly and non-identifiable narratives to determine risk of re-identification. This identified that the only leaked full name was of Indian origin, leading to a follow-up analysis to determine if there was a problem with under-serving certain types of names.<br>- Systematic case-level review of all false positives, with diagnostic tags identifying masked tokens containing clinically relevant information.<br>- Scientific publication included examples of true positives, false negatives and false positives with scrambled personal identifiers illustrating the abilities and mistakes of the method. |

## Performance evaluation of human-AI teams

Many AI systems are designed for intelligence augmentation. This means they are intended to support and enhance human decision-making rather than replace it or fully automate a task. In this context, evaluation could also consider the performance and dynamics of the human-AI team.

One important aspect is the assessment of team-level outcomes rather than evaluating the AI system in isolation. Metrics such as recall, precision, can still be informative. Additionally, decision efficiency, including the time and resources required may also be factored in. However, experimental designs must consider the challenge that the same dataset cannot typically be used to simultaneously evaluate both human-only and human plus AI performance. Once a human has reviewed a case, their subsequent decisions may be influenced by prior exposure, which introduces bias into the comparison.

Another important aspect for evaluation is how well the AI system integrates into the human workflow. Measures such as 'decision concordance' and 'override rate' by a human-in-the-loop can reveal the degree of alignment between AI recommendations and human judgment. The format of the AI output, whether visual or textual, can significantly influence usability and trust (27). Evaluation methods that focus on the user experience, such as usability testing, are useful for understanding how users adapt to and benefit from the system over time.

Trust in AI systems is affected by their opacity. Users are more likely to trust systems whose reasoning they can understand (14). Evaluating AI-human teams requires frameworks that go beyond individual technology acceptance, focusing instead on how AI systems integrate into collaborative workflows. Human Factors Engineering offers an approach for this purpose. As explained by Sujan et al. (2022), Human Factors Engineering principles—such as automation bias, explanation, trust and human-AI teaming are essential for designing and assessing AI systems that support effective human-AI collaboration (28).

# Checklist

*Table 1*. Key questions to consider in procurement or development of AI models for rare-event recognition to ensure validity, robustness and fairness.

| CONSIDERATION | KEY QUESTIONS |
|---|---|
| **Test sets** | How well does the nature of information and scope of available test sets align with the intended use case? Are they large and diverse enough? Are there enough positive controls and are they representative of events in scope for the intended use case? Are there enough negative controls and are they representative of events out of scope for the intended use case? |
| **Annotation process** | What are the criteria for positive and negative controls and were adequate measures taken to ensure and assess the quality and consistency of annotations? How are edge cases managed? |
| **Metrics** | What performance metrics are used and are they relevant to the intended use case? Together, do they capture enough aspects of performance relevant to the intended use case (e.g. false positives, false negatives, stability)? |
| **Recall** | If recall is evaluated, do test sets reflect the full spectrum of positive controls (types, difficulty, …)? If there is enrichment with positive controls, is this adequately accounted for? |
| **Precision** | If precision is evaluated, what is the prevalence of positive controls in the test set and how does that correspond to the intended use case? If there is enrichment with positive controls, is this adequately accounted for? |

| Specificity | If specificity is evaluated, is it high enough for the desired operating point and has it been reliably estimated? |
|---|---|
| Decision thresholds | What decision thresholds are considered in the performance evaluation, and do they align with the intended use case and relative costs of errors? |
| Benchmarks | Have comparisons been made to relevant benchmark methods and if so, are adequate measures taken to ensure that benchmark methods perform to the best of their ability? Are relevant benchmark test sets available for intended use case, and, if so, is the performance of the proposed AI model on these sufficient? |
| Robustness | Does the AI model perform well enough under varying conditions and across relevant subsets. Are sufficient measures in place to identify and respond to data, model, or performance drifts? |
| Non-triviality | What is the nature of true positives identified by the AI model - do they reflect a capacity to detect non-trivial events of interest? |
| Types of errors | What is the nature of false positives and false negatives? Are they understandable and acceptable for the intended use case, or do they raise concerns regarding the validity or fairness of the AI model? |
| Human-AI interaction | What is the intended human-AI interaction and has it been appropriately accounted for in performance evaluation? |

# Discussion

The cost in time and effort to develop artificial intelligence systems is rapidly decreasing, with the advent of pre-trained generative language models that offer unprecedented versatility and can be deployed without fine-tuning (1). In contrast, rigorous performance evaluation remains resource intensive. This imbalance creates a risk that organizations devote less time to understanding limitations and sources of error, while being tempted to cut corners in appraisal. The danger is not only wasted resources but also negative impact on stakeholders and eroded trust if systems underperform in practice. A risk-based approach is essential: bold experimentation may be appropriate where costs of errors are contained and possible to reverse, whereas careful evaluation remains indispensable in settings where costs are substantial or irreversible. Our framework is intended as a backbone that can be adapted to context, rather than a rigid one-size-fits-all prescription.

The principles described here are largely method-agnostic and apply to rare event-recognition in general. However, the extent of performance evaluation should account for the ambiguity of the task, and the opacity, adaptiveness, scope, and autonomy of the AI system. Prevalence-aware statistical evaluation, transparency in test set design and annotation, and the use of our proposed structured case-level examination to complement summary metrics are broadly relevant. So is performance assessment across subsets of data and other approaches to robustness analysis. It is worth noting that AI models that produce rankings of events become binary classifiers after thresholding. Similarly, while methods for mapping free text verbatims to standard terminologies and other multiclass classification problems are not rare event-recognition per se, prevalence-aware statistical evaluation can still be relevant. For example, overall evaluation of a method for mapping free text to adverse event terms may be dominated by more common terms, with strong general performance hiding a failure to effectively recognize rare adverse events that can be of crucial importance in pharmacovigilance. It can therefore be helpful to perform complementary performance analyses considering a set of binary event-recognition classification tasks, using the principles outlined here.

While the principles underpinning SCLE have been derived from real-world experience, the more systematic framework proposed here needs testing and refinement under real-world conditions. It should also be noted that by design, case-level examination is vulnerable to sampling variability, and observed patterns may require confirmation in follow-up analyses. These are not reasons to forgo this aspect of appraisal but rather to motivate careful design, transparent reporting, and ongoing methodological refinement.

Looking ahead, two developments deserve special attention. First, there is growing interest in using AI itself to support performance evaluation, for example by employing generative LLMs to annotate test sets or review output from (simpler) AI models (29). The use of LLMs as a judge may help reduce the resource burden of evaluation and allow performance evaluation at entirely new scales (30), but their validity must be ensured and demonstrated via human calibration. They may be particularly valuable in continual performance monitoring after deployment, since human specialists may be more difficult to engage at this stage. Second, generative applications are becoming more widespread where AI produces free text rather than numerical or categorical output (31). It is possible that some of the systems that perform rare event-recognition will be replaced by other modalities. For example, redaction of person names could be addressed as a text editing task instead of as binary token classification. This would require a different approach to statistical evaluation, but SCLE would still be relevant with different categories for the stratified random sampling. SCLE may also be valuable if relying on an LLM-as-a-judge for performance evaluation. Many workflows involving human operators will likely continue to rely on dichotomous decisions (spam/no spam, human review or not). The same is true of use cases where event recognition is the basis for subsequent analyses. However, processes that evolve to a more fluid back-and-forth exchange between human operators and AI (32), will require a different evaluation paradigm, assessing real-world decision-making by human-AI teams, accounting for user experience, cognitive ergonomics, and decision efficiency.

As AI technologies and practices evolve, so too must the standards for their evaluation. What remains constant is the need for appraisal that balances efficiency with rigor, enabling organizations to harness the benefits of AI for rare-event recognition while safeguarding validity, robustness, and fairness.

# Appendix A: Intended use and test sets for use cases

**Example: VigiBase pregnancy algorithm**
- **Intended use**: Identifying adverse event reports related to pregnancy for individual case review or subsequent statistical analyses.
- **Scope:** All deduplicated VigiBase reports up to January 2023.
- **Out of scope:** The method does not analyze unstructured free text fields, which may contain relevant pregnancy information. Paternal exposures are not considered.
- **Reference standard:** Positive controls: reports with evidence suggesting they refer to cases where drug was administered to a pregnant person, before or during pregnancy/labour, with or without adverse event in the pregnant
- person or the foetus/prenatally exposed child. Negative controls: reports without such evidence, including reports where drugs were administered to a father, drug administered to the pregnant person postpartum or during lactation (unless administered also during pregnancy), drug administered to the child after birth, drug administered to non-pregnant person.
- **Prevalence of positive controls**: 3.2% of the reports in the downsampled data set of VigiBase were annotated as related to pregnancy. Based on pre-study results, the prevalence of pregnancy cases in VigiBase has been estimated to be around 1%.
- **Annotation process:** The annotations were performed by a team of experienced pharmacovigilance assessors following a predefined guideline. Inter-annotator agreement was quantified using Fleiss' kappa to be 83%, indicating a very good level of agreement.
- **Ambiguity:** Borderline cases, where pregnancy status was unclear or inconsistently documented, posed challenges for annotators, contributing to uncertainty in the reference standard. Ambiguity in the annotation process primarily stemmed from limitations in the structured data available in case reports. In many instances, pregnancy-related information was present only in free-text narratives, which are not processed by the algorithm (and missing altogether from many reports due to restrictions in how these data is shared).

**Example: vigiMatch duplicate detection method**
- **Intended use**: Highlighting suspected duplicate reports for subsequent human review and/or removal of suspected duplicates prior to statistical analyses of adverse event reporting data.
- **Scope**: VigiBase adverse event reports (27 million for medicines and 1.8 million for vaccines) available in January 2023.

- **Out of scope**: Nearly 5 million reports related to COVID-19 vaccines were excluded (due to their overrepresentation and atypical reporting patterns)
- **Reference standard**: Positive controls: Possible/likely duplicates; Negative controls: Non-duplicates (model-specific precision tests additionally included a second subcategory of negative controls corresponding to Otherwise related reports).
- **Prevalence of positive controls**: The prevalence of duplicate adverse event reports is not known and may vary. A recent study by the US FDA found 20% of the reports in the FAERS database to be suspected duplicates (33). Since the analysis of duplicates is at the level of report pairs, if 20% of the reports in a database of 40 million reports have a (single) duplicate, the prevalence of duplicates among all possible pairs of reports is expected to be 1 in 200 million ($0,2*40*10^6 / (40*10^6)^2$).
- **Annotation process**: Recall study based on already annotated data from four different national regulatory authorities (for which annotation process is not well-described). Annotations for model-specific precision tests followed annotation guideline developed for study and included as supplementary material in Barrett et al (2025) (23). All pairs in the precision tests were doubly annotated to determine inter-annotator agreement.
- **Ambiguity**: Many reports pairs in VigiBase are not possible for human operators to to classify with confidence as duplicates or non-duplicates because of the limited available information on reports. Inter-annotator agreement was estimated to a Cohen-Kappa score of 0.67, which suggests moderate to substantial agreement between human annotators, reflecting this ambiguity.

**Example: UMC redaction of narratives method**
- **Intended use**: Highlighting possible person names for human review and redaction or automated redaction of possible person names in case narratives as a privacy-preserving measure.
- **Scope:** Yellow Card data, with case narratives written in English (UK) as they arrive at the pharmacovigilance organisation
- **Out of scope:** Excluding narratives derived from electronic health records as they were known to have lower expected prevalence of person names, and therefore be of less interest for redaction.
- **Reference standard:** Positive controls: NAME tokens; negative controls: NON-NAME tokens
- **Prevalence of positive controls:** 71 out 5,042 randomly sampled narratives (1.4%) included at least one person name, and in total 179 tokens were annotated as NAME compared to 263,272 as NON-NAME (0.07%).
- **Annotation process:** De novo annotations performed by four different annotators according to annotation guideline developed for purpose of study (see Meldau et al supplementary material). Lack of double annotations and of study of intra-annotator agreement acknowledged as a limitation.
- **Ambiguity:** There were difficult cases involving short tokens that could not be unambiguously classified (despite consultation with native English speakers from the UK, one of whom is a medical doctor), illustrating the fundamental difficulty

> and ambiguity of the task. When not possible to determine with confidence, tokens were annotated as NAME (aligned with false negatives being considered more costly than false positives).

# Acknowledgments

GNN would like to acknowledge the other members of the CIOMS XIV working group on Artificial Intelligence in Pharmacovigilance and its Editorial Team for inspiring and challenging discussions on the topic. Earlier version of the considerations for prevalence-aware statistical performance evaluation was presented to the CIOMS XIV working group and received critical and helpful feedback. The views expressed here are those of the authors and not necessarily those of the Uppsala Monitoring Centre or any other organization.

## Ethics declarations

**Conflict of Interest**

The authors declare no conflict of interest related to this work.

**Ethics Approval**

Not applicable.

**Availability of Data and Material**

Not applicable

**Code Availability**

Not applicable.